\pdfoutput=1
\documentclass[11pt]{article}

\usepackage[preprint]{acl}

\usepackage{times}
\usepackage{latexsym}

\usepackage[T1]{fontenc}

\usepackage[utf8]{inputenc}

\usepackage{microtype}

\usepackage{inconsolata}

\usepackage{graphicx}
\usepackage{algorithm}
\usepackage{algorithmic}

\usepackage{newfloat}
\usepackage{listings}
\usepackage{booktabs}
\usepackage{multirow}
\usepackage{amsmath}
\usepackage{amssymb}
\usepackage{array}
\usepackage{tabularx}
\usepackage[table]{xcolor}

\definecolor{fullresgray}{gray}{0.91}
\newcommand{\gray}[1]{\textcolor[gray]{0.55}{#1}}
\newcommand{\dgray}[1]{\textcolor[gray]{0.35}{#1}}
\newcommand{\second}[1]{\underline{#1}}

\newcolumntype{Y}{>{\centering\arraybackslash}X}

\title{Visual Tokens on Demand: Visual Budgeting for Scalable and Efficient \\ Multi-Image Reasoning in Multimodal LLMs}

\author{
  \textbf{Aimon Rahman},
  \textbf{Joymallya Chakraborty},
  \textbf{Vidit Bansal},
  \textbf{Abhishek Dan}
\\
  Amazon
\\
  \small{
    \texttt{\{aimonrah, joymally, bansalv, abhidan\}@amazon.com}
  }
}

\begin{document}
\maketitle
\begin{abstract}
Multimodal large language models (MLLMs) have advanced rapidly, but most remain optimized for single-image inputs. Multi-image reasoning is costly because each image adds many visual tokens, increasing encoding and attention demands. Existing techniques that address the above issue, including visual token pruning, token merging, visual token compression, learned resampling, hierarchical compression, retrieval-augmented visual reasoning, and memory-based architectures, have largely been designed for single-image scenarios, where redundancy is primarily spatial and local. Multi-image reasoning introduces additional challenges: redundancy may occur across images, task-relevant evidence may be sparse or distributed, and aggressive per-image compression can remove details needed for comparison, counting, temporal reasoning, retrieval, or evidence grounding. Moreover, existing token-reduction methods are often either (i) \textit{training-free} but inference-inefficient, typically relying on accessing attention matrices to get token-importance that may be impractical to access with memory-efficient attention implementations such as FlashAttention, or (ii) \textit{training-dependent}, requiring the base MLLM to be retrained or adapted to a new compressed-token paradigm. In this work, we study scalable multi-image reasoning and understanding without retraining the base model and without accessing attention matrices at inference time. To this end, we propose \textbf{\textit{mmVB}} (\textbf{\underline{m}}ulti-\textbf{\underline{m}}odal \textbf{\underline{V}}isual \textbf{\underline{B}}udgeter), a lightweight \textit{plug-and-play} module that assigns query-conditioned per-image token budgets via cross-image context, adding only 1.6M trainable parameters while reducing visual input tokens by 9$\times$. This reduction yields 5$\times$ faster time-to-first-token and 7$\times$ fewer prefill FLOPs, while maintaining 94.4\% of full-resolution accuracy across eight multi-image reasoning benchmarks (geometric mean of per-benchmark ratios). These results show that input-aware visual scaling can substantially improve efficiency while preserving multi-image reasoning performance.
\end{abstract}
\section{Introduction}

\begin{figure*}[!t]
    \centering
    \includegraphics[width=\linewidth]{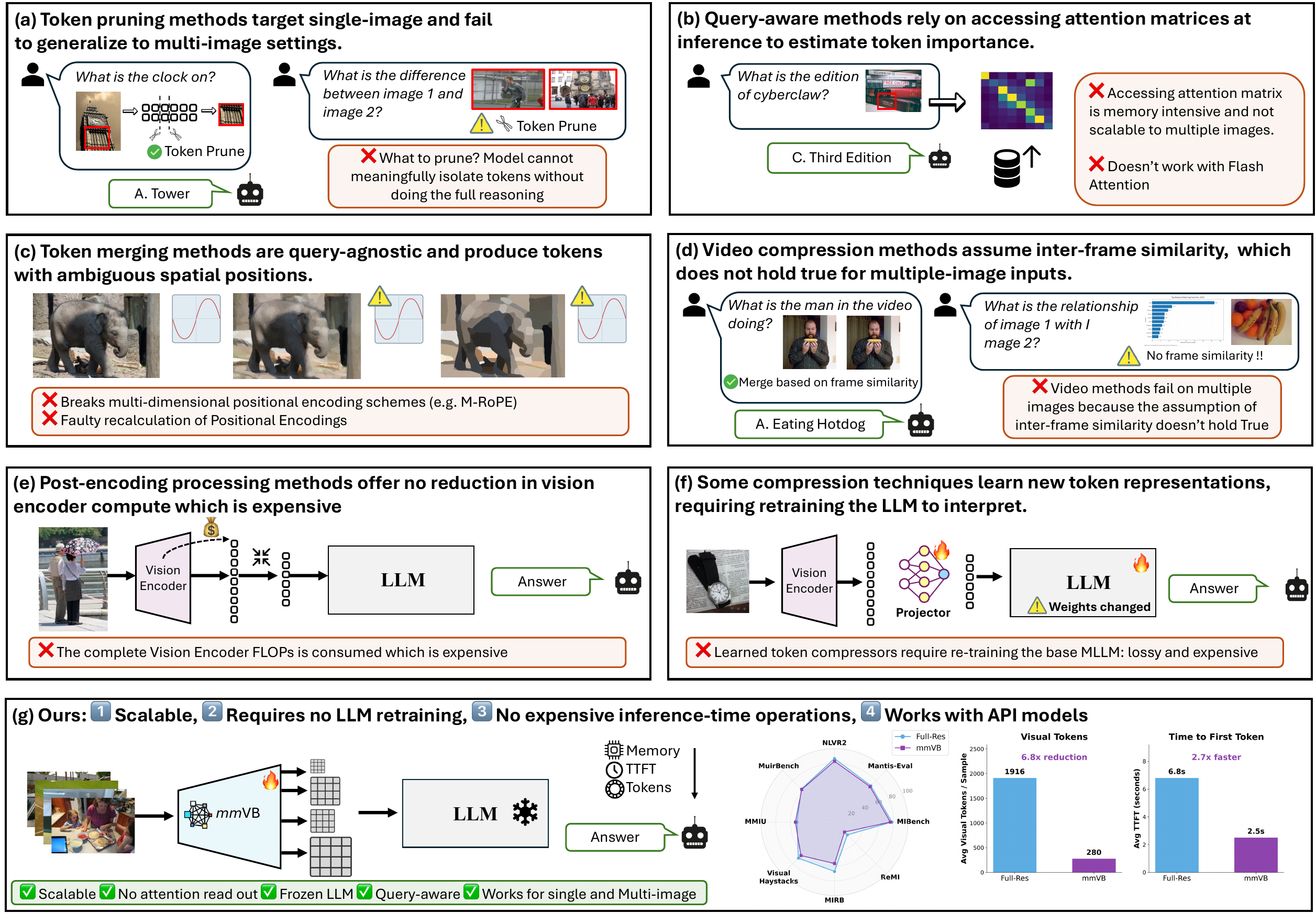}
    \caption{\textbf{Introduction and Motivation.} Existing multi-image methods scale poorly: dense attention breaks FlashAttention and excludes API-only models, token merging disrupts positional encodings and cross-image reasoning, and video-style compression assumes frame similarity. Our pre-encoding predictor instead assigns query-conditioned token budgets using only pixels, enabling compatibility with any MLLM, including closed models.}
    \label{fig:intro}
\end{figure*}
Although current multimodal large language models (MLLMs) can accept multiple images as input, the problem of scaling them efficiently to many-image settings remains underexplored. Multi-image understanding and reasoning is inherently challenging: the model must identify relevant visual evidence, compare information across images, and aggregate details that may be distributed over the input set. Scaling this capability further increases the difficulty, as every additional image contributes visual tokens and raises memory, latency, and compute costs. This limitation is especially important in real-world applications. For instance, a catalog-oriented LLM may need to process multiple product images to understand item attributes, compare variants, and answer user queries, while operating within a constrained inference budget.

Visual token redundancy has been widely studied, with many works showing that multimodal language models can operate effectively, and in some cases even better, with fewer visual tokens. Existing approaches typically reduce tokens through pruning, merging, or learned compression, either in a training-free manner or by training additional modules. However, applying these methods to scalable multi-image reasoning presents three key challenges: \textbf{i) Limited systems scalability.} Many training-free methods rely on accessing attention matrices, often across multiple layers, to estimate token importance. This can be memory-intensive and difficult to integrate with optimized inference kernels such as FlashAttention, which avoid materializing dense attention maps. Production systems often depend on such kernels for speed and memory efficiency, meaning that methods requiring explicit access to dense attention matrices may be impractical at scale. Token merging can also disrupt positional structure, and correcting the resulting positional encoding mismatch often requires additional architectural changes. \textbf{ii) Insufficient study of multi-image reasoning.} Most token-reduction methods are designed for single-image inputs, where task-relevant regions are often spatially localized. For example, when answering \textit{“What is the species of the animal in this image?”}, attention may concentrate on the animal region, making pruning or merging relatively effective, often even from early layers. In the multi-image reasoning case, questions such as \textit{“What is the difference between image 1 and image 2?”} may require preserving fine-grained details across both images until later stages of reasoning. In such cases, aggressive early pruning can remove evidence that only becomes important after cross-image comparison. \textbf{iii) Training dependence and distribution shift.} Some approaches require training adapters, compressed tokens, or modified visual-language interfaces, which can change the input distribution seen by the language model or require adapting its weights. This is problematic when the goal is to scale existing MLLMs without retraining the base model, not overfitting to a particular dataset, or altering the model’s original reasoning behavior. Moreover, recent evidence shows that simple image downsampling consistently outperforms sophisticated post-encoding compression methods at matched token budgets~\cite{liao2026we}, questioning the fundamental value of operating after the vision encoder. Together, these limitations suggest the need for a token-reduction strategy that is scalable, compatible with efficient inference, applicable to multi-image inputs, and usable with frozen multimodal language models.

A natural solution is to reduce the number of visual tokens before language-model processing, and ideally even before the visual encoder incurs the full cost of processing all image content. Under a constrained-budget setting, we formulate the problem as estimating the minimum relative visual scale needed for the downstream language model to answer a visual question correctly. The scale estimator must itself be lightweight and avoid introducing additional memory overhead, since the goal is to enable efficient reasoning at scale. To this end, we propose \textbf{mmVB}, a lightweight plug-and-play module that assigns input-dependent visual budgets across images conditioned on both the image set and the question. Rather than applying a fixed resolution or token budget to every image, Visual Budgeter estimates how much visual detail each image requires for the downstream reasoning task. The motivation of this work is illustrated in Figure \ref{fig:intro}. Our main contributions are:
\begin{itemize}
\item[-] We formulate scalable multi-image reasoning as a visual budget allocation problem, where the goal is to estimate the minimum image scale needed for accurate downstream reasoning under a constrained compute budget. We propose \textbf{mmVB}, a scalable, lightweight \textit{plug-and-play} module with only 1.6M trainable parameters that assigns question-conditioned visual budgets across images without retraining the base MLLM.
\item[-] This Visual Budgeter reduces visual input tokens by 9$\times$, yielding 5$\times$ faster time-to-first-token and 7$\times$ fewer prefill FLOPs.
\item[-]We evaluate Visual Budgeter across eight multi-image reasoning benchmarks and show that it preserves 94.4\% of full-resolution accuracy, demonstrating that input-aware visual scaling can substantially improve efficiency while maintaining strong multi-image reasoning performance.
\end{itemize}
\section{Proposed \textit{mm}VB Method}
\begin{figure}[!t]
    \centering
    \includegraphics[width=\linewidth]{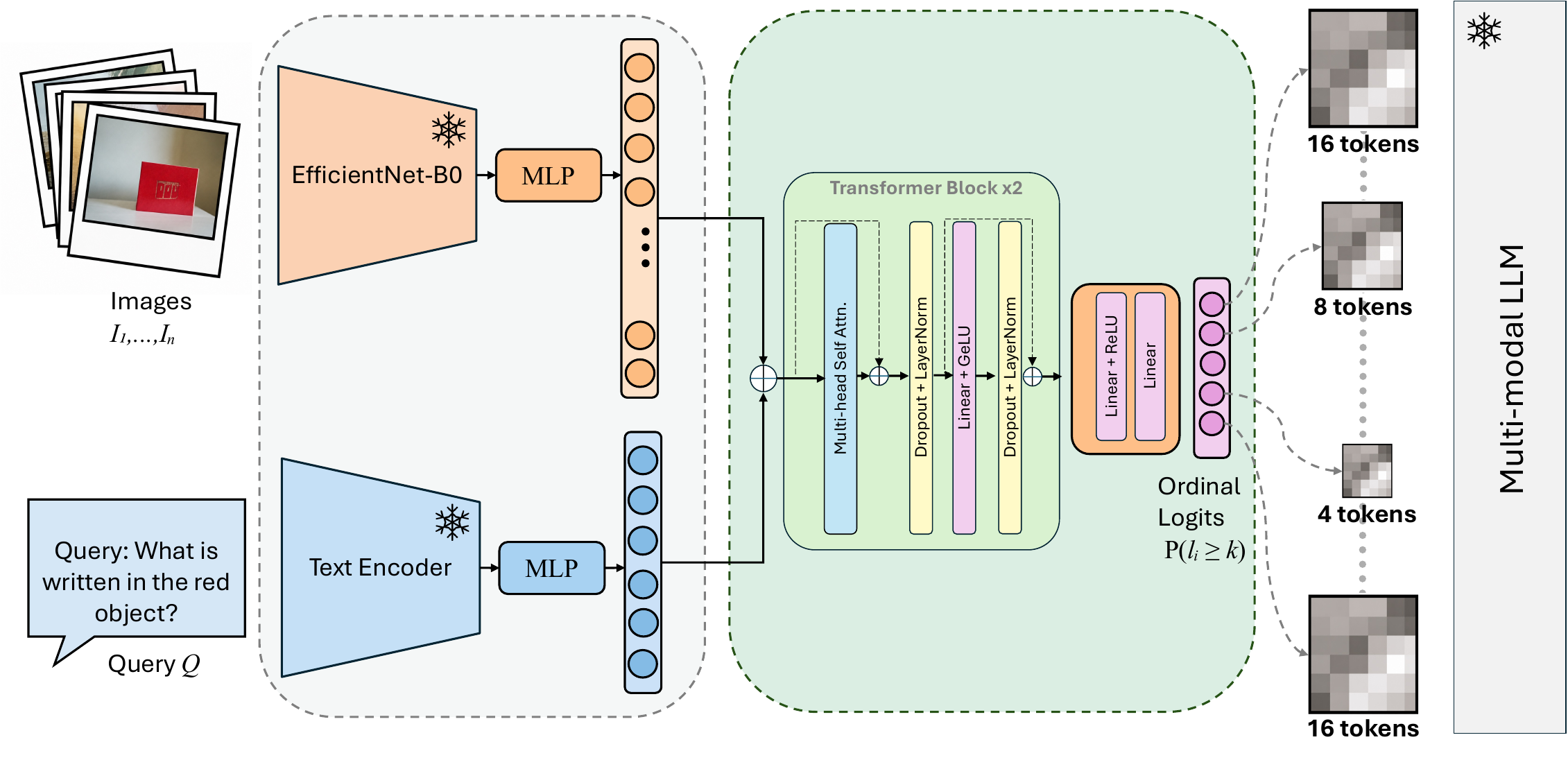}
    \caption{\textbf{mmVB architecture.} Frozen EfficientNet-B0 encodes $N$ images, while frozen LLM embeddings encode the query through mean pooling and a 64-dimensional bottleneck. A two-layer, four-head Transformer processes all $N{+}1$ tokens, after which a shared MLP maps each image token to four ordinal logits.}
    \label{fig:archi}
\end{figure}
\subsection{Architecture}

Given $N$ images $\{I_1,\ldots,I_N\}$ of arbitrary resolutions and a text query $Q$, the Visual Budgeter assigns each image a token-allocation level $\ell_i \in \{0,1,2,3,4\}$ that approximates the minimum number of visual tokens the MLLM needs to answer correctly. In doing so, the budgeter also learns each image's relative relevance to the query. The full architecture is shown in Fig.~\ref{fig:archi}.

\paragraph{Encoding and Context Fusion.}
Each image is encoded by a frozen EfficientNet-B0~\cite{tan2019efficientnet} and linearly projected to $d=256$. The query is embedded with the frozen MLLM embedding layer, mean-pooled, and mapped to the same space through a 64-dimensional bottleneck. The query token and $N$ image tokens are then processed by a two-layer, four-head Transformer encoder, which performs query-conditioned, cross-image reasoning about each image's required detail and relative importance.

\paragraph{Resolution Head and Inference.}
A shared MLP ($256 \rightarrow 64 \rightarrow 4$) maps each image representation to four ordinal logits with $P(\ell_i \geq k)=\sigma(z_{i,k})$ for $k \in \{1,2,3,4\}$, and the predicted level is $\hat{\ell}_i=\sum_{k=1}^{4}\mathbb{I}[\sigma(z_{i,k})>0.5]$. Each image is then resized to its predicted level (never upscaled) before being passed to the MLLM. The budgeter has about 1.6M trainable parameters and adds roughly 2\,ms of overhead. Layer specifications, parameter counts, design choices, and an architecture ablation are given in Appendix~\ref{app:arch}.

\subsection{Training}
\paragraph{Loss Function.}
We train the model using ordinal binary cross-entropy. For a ground-truth level $\ell_i^* \in {0,1,2,3,4}$, the cumulative targets are
\begin{equation}
y_{i,k}=\mathbf{1}\left[\ell_i^* \geq k\right],
\qquad k \in {1,2,3,4}.
\end{equation}
Given predicted logits $z_{i,k}$, the threshold probabilities are
\begin{equation}
p_{i,k}=\sigma\left(z_{i,k}\right),
\end{equation}
where $p_{i,k}$ estimates $P(\ell_i \geq k)$. The loss for image $i$ is
\begin{equation}
\mathcal{L}_{i}^{\mathrm{ord}} =
-\sum_{k=1}^{4}
\left[
y_{i,k}\log\left(p_{i,k}\right)
+
\left(1-y_{i,k}\right)
\log\left(1-p_{i,k}\right)
\right].
\end{equation}
This formulation preserves the ordering of the levels: predictions closer to $\ell_i^*$ violate fewer thresholds and incur a smaller loss. For example, when $\ell_i^*=3$, predicting $2$ violates one threshold, while predicting $0$ violates three. Full training details (optimizer, learning rate, schedule, and batch configuration) are provided in the supplementary material.

\subsection{Training-data Generation}
We address image interactions using a three-round procedure: independent ascending search, joint verification, and iterative relaxation (Algorithm~\ref{alg:gt_gen}). For each image, we sweep Levels~$0$--$4$ and select the first level that yields a correct MLLM response, without assuming monotonic accuracy across resolutions. Single-image samples require at most five forward passes plus a pre-check, while multi-image samples cost $O(5N+KN)$ in the worst case. With $K=5$, joint verification succeeds without relaxation for most samples, so the average cost remains close to independent search. Label generation is performed once offline, and the resulting labels are reused across training runs and hyperparameter sweeps. A text-only ablation assigns nearly uniform levels, showing that meaningful allocation requires modeling the query$\times$image-content interaction through multi-resolution MLLM evaluation.

\begin{algorithm}[t]
\caption{Training Label Generation}
\label{alg:gt_gen}
\footnotesize
\setlength{\algorithmicindent}{1.5em}
\renewcommand{\algorithmiccomment}[1]{\hfill\textcolor[gray]{0.5}{\(\triangleright\) #1}}
\begin{algorithmic}[1]
\REQUIRE Sample $(Q, \{I_1, \ldots, I_N\}, A)$, MLLM $\mathcal{M}$
\ENSURE Resolution levels $\ell_1, \ldots, \ell_N$
\STATE \textbf{Pre-check:} If $\mathcal{M}(Q, I_1, \ldots, I_N) \neq A$, discard sample

\STATE \textbf{Round 1: Independent Ascending Search}
\FOR{$i = 1$ to $N$}
    \FOR{$\ell = 0$ to $4$}
        \IF{$\mathcal{M}(Q, I_1, \ldots, R_{\ell}(I_i), \ldots, I_N) = A$}
            \STATE $\ell_i \leftarrow \ell$; \textbf{break} \COMMENT{First sufficient level}
        \ENDIF
    \ENDFOR
\ENDFOR
\STATE \textbf{Round 2: Joint Verification}
\IF{$\mathcal{M}(Q, R_{\ell_1}(I_1), \ldots, R_{\ell_N}(I_N)) = A$}
    \RETURN $\ell_1, \ldots, \ell_N$
\ENDIF
\STATE \textbf{Round 3: Iterative Relaxation}
\FOR{$k = 1$ to $K$}
    \FOR{$i = 1$ to $N$}
        \IF{$\ell_i < 4$}
            \STATE $\ell_i' \leftarrow \ell_i + 1$
            \IF{$\mathcal{M}(Q, \ldots, R_{\ell_i'}(I_i), \ldots) = A$}
                \RETURN $\ell_1, \ldots, \ell_i', \ldots, \ell_N$
            \ENDIF
        \ENDIF
    \ENDFOR
    \STATE $\ell_i \leftarrow \min(\ell_i + 1, 4)$ for all $i$ \COMMENT{Bump all}
    \IF{$\mathcal{M}(Q, R_{\ell_1}(I_1), \ldots, R_{\ell_N}(I_N)) = A$}
        \RETURN $\ell_1, \ldots, \ell_N$
    \ENDIF
\ENDFOR
\RETURN $(4, \ldots, 4)$ \COMMENT{Safe: pre-check verified full res.}
\end{algorithmic}
\end{algorithm}

\paragraph{Label Interpretation.}
The labels produced by Algorithm~\ref{alg:gt_gen} are \emph{sufficient feasible allocations}, not provably minimal ones. The ascending sweep finds the first working level per image without assuming that correctness is monotonic in resolution; even if a higher level were to fail due to MLLM stochasticity, the returned level is verified correct. Note that the pre-check (line~1) guarantees that full resolution produces the correct answer; consequently, both the Round~1 sweep (which tests each image at level~4 while others remain at full resolution) and the line~29 fallback to $(4,\ldots,4)$ are guaranteed to be valid. No sample that fails at full resolution enters the label generation pipeline. For multi-image samples, the greedy relaxation in Round~3 searches single-image increments and then uniform increments, which may not discover the cheapest subset combination (e.g., two images needing simultaneous promotion). This means the returned allocation can over-allocate relative to the true joint minimum. Crucially, this bias is \emph{conservative}: labels may assign more tokens than strictly necessary but never fewer, since every returned vector is verified to produce the correct answer. For training, conservative labels are preferable to aggressive ones they train the predictor toward accuracy preservation rather than over-compression. The ordinal loss further tolerates label noise: a label of Level~3 when the true minimum is Level~2 incurs only one incorrect threshold, producing a mild penalty rather than a catastrophic gradient signal.
\begin{table*}[!h]
\centering
\small

\setlength{\tabcolsep}{9pt}
\renewcommand{\arraystretch}{1.02}

\begin{tabularx}{\textwidth}{
    @{}
    >{\raggedright\arraybackslash}p{3.25cm}
    *{8}{Y}
    @{}
}
\toprule

\textbf{Method}
& \textbf{Muir}
& \textbf{MIB}
& \textbf{Mantis}
& \textbf{NLVR2}
& \textbf{VHay}
& \textbf{MMIU}
& \textbf{MIRB}
& \textbf{ReMI} \\

\textit{Samples}
& 2,600
& 5,000
& 217
& 6,967
& 20,100
& 93,584
& 1,124
& 2,600 \\

\midrule

\multicolumn{9}{@{}l}{
    \textbf{Accuracy (\%) $\uparrow$}
} \\
\addlinespace[1pt]

\rowcolor{fullresgray}
Full Res.
& 63.6
& 78.1
& 69.9
& 86.5
& 69.5
& 50.9
& 67.2
& 24.7 \\

VisionZip \textit{(CVPR'25)}
& \textbf{63.6}
& 72.8
& \textbf{69.6}
& 59.3
& \textbf{69.6}
& 48.3
& 48.7
& 13.8 \\

ToFu \textit{(arXiv)}
& 42.6
& 71.2
& 56.7
& 60.7
& 14.6
& 11.2
& 42.1
& 18.6 \\

VisionTrim \textit{(ICLR'26)}
& \textbf{63.6}
& \textbf{80.4}
& 67.7
& \second{80.6}
& \second{64.2}
& \second{52.5}
& \textbf{59.6}
& \textbf{19.3} \\

ZooPrune \textit{(CVPR'26)}
& 57.0
& 69.9
& 65.4
& 78.3
& 63.4
& 49.6
& 49.5
& 16.8 \\

mmVB \textit{(Ours)}
& \second{62.9}
& \second{77.9}
& \second{68.2}
& \textbf{82.8}
& \textbf{69.6}
& \textbf{53.2}
& \second{56.5}
& \second{19.2} \\

\addlinespace[2pt]
\midrule
\addlinespace[1pt]

\multicolumn{9}{@{}l}{
    \textbf{Visual tokens/sample $\downarrow$}
} \\
\addlinespace[1pt]

\rowcolor{fullresgray}
Full Res.
& 3,236
& 1,237
& 2,119
& 1,649
& 2,187
& 2,980
& 1,756
& 2,573 \\

VisionZip \textit{(CVPR'25)}
& 2,265
& 462
& 2,024
& 199
& 1,836
& 712
& 891
& 286 \\

ToFu \textit{(arXiv)}
& 1,731
& 1,122
& 2,050
& 1,171
& 1,557
& 522
& 1,391
& 2,177 \\

VisionTrim \textit{(ICLR'26)}
& 3,531
& 370
& 634
& 494
& 1,165
& 1,532
& 865
& 770 \\

ZooPrune \textit{(CVPR'26)}
& \textbf{356}
& \textbf{136}
& \second{182}
& \second{95}
& \textbf{239}
& \textbf{300}
& \textbf{159}
& \textbf{99} \\

mmVB \textit{(Ours)}
& \second{389}
& \second{248}
& \textbf{118}
& \textbf{89}
& \second{594}
& \second{453}
& \second{193}
& \second{155} \\

\addlinespace[2pt]
\midrule
\addlinespace[1pt]

\multicolumn{9}{@{}l}{
    \textbf{Prefill computation (TFLOP) $\downarrow$}
} \\
\addlinespace[1pt]

\rowcolor{fullresgray}
Full Res.
& 22,300
& 18,507
& 1,170
& 48,969
& 10,828
& 138,050
& 6,049
& 44,360 \\

VisionZip \textit{(CVPR'25)}
& 20,835
& 12,329
& 1,282
& 14,997
& 9,523
& 79,901
& 4,898
& 8,118 \\

ToFu \textit{(arXiv)}
& 9,189
& 13,623
& 1,129
& 24,054
& \textbf{1,687}
& \textbf{4,358}
& \second{3,339}
& 33,084 \\

VisionTrim \textit{(ICLR'26)}
& 83,339
& 16,720
& 1,052
& 27,379
& 5,600
& 120,409
& 13,686
& 15,273 \\

ZooPrune \textit{(CVPR'26)}
& \second{9,148}
& \second{6,441}
& \second{422}
& \second{9,097}
& 42,408
& 40,554
& 7,039
& \second{3,534} \\

mmVB \textit{(Ours)}
& \textbf{2,830}
& \textbf{4,698}
& \textbf{129}
& \textbf{3,146}
& \second{2,761}
& \second{11,429}
& \textbf{2,265}
& \textbf{3,042} \\

\addlinespace[2pt]
\midrule
\addlinespace[1pt]

\multicolumn{9}{@{}l}{
    \textbf{Time to first token (TTFT, seconds) $\downarrow$}
} \\
\addlinespace[1pt]

\rowcolor{fullresgray}
Full Res.
& 3.5
& 1.6
& 2.2
& 2.8
& 11.8
& 7.9
& 2.5
& 22.2 \\

VisionZip \textit{(CVPR'25)}
& \second{1.1}
& 1.3
& 2.6
& 1.6
& \second{0.8}
& 1.1
& 2.3
& 17.9 \\

ToFu \textit{(arXiv)}
& 7.3
& 5.5
& 9.9
& 6.5
& 27.3
& 15.5
& 6.8
& 22.5 \\

VisionTrim \textit{(ICLR'26)}
& 1.2
& \textbf{0.6}
& 0.7
& \second{0.6}
& 1.2
& 1.2
& \second{1.0}
& \textbf{11.4} \\

ZooPrune \textit{(CVPR'26)}
& \textbf{0.5}
& \second{0.7}
& \textbf{0.5}
& \textbf{0.5}
& \textbf{0.6}
& \second{0.8}
& \textbf{0.7}
& \second{11.6} \\

mmVB \textit{(Ours)}
& \textbf{0.5}
& \second{0.7}
& \second{0.6}
& \second{0.6}
& \textbf{0.6}
& \textbf{0.6}
& 1.3
& 15.2 \\

\addlinespace[2pt]
\midrule
\addlinespace[1pt]

\multicolumn{9}{@{}l}{
    \textbf{Harmonic mean (HM) $\uparrow$}
} \\
\addlinespace[1pt]

VisionZip \textit{(CVPR'25)}
& 0.46
& 0.75
& 0.09
& 0.77
& 0.28
& \second{0.84}
& 0.59
& 0.69 \\

ToFu \textit{(arXiv)}
& 0.55
& 0.17
& 0.06
& 0.41
& 0.24
& 0.35
& 0.31
& 0.26 \\

VisionTrim \textit{(ICLR'26)}
& 0.00
& \second{0.83}
& 0.81
& 0.80
& 0.62
& 0.66
& 0.65
& 0.74 \\

ZooPrune \textit{(CVPR'26)}
& \second{0.89}
& \textbf{0.89}
& \second{0.92}
& \second{0.92}
& \textbf{0.90}
& \textbf{0.94}
& \second{0.81}
& \second{0.80} \\

mmVB \textit{(Ours)}
& \textbf{0.93}
& \textbf{0.89}
& \textbf{0.96}
& \textbf{0.95}
& \second{0.82}
& \textbf{0.94}
& \textbf{0.86}
& \textbf{0.85} \\

\bottomrule
\end{tabularx}

\vspace{4pt}

\begin{minipage}{0.98\textwidth}
\small
\raggedright
\textit{Notes.}
Muir = MuirBench;
MIB = MIBench;
Mantis = Mantis-Eval;
VHay = Visual Haystacks.
VisionTrim uses DVTS + TGVC post-encoding token selection with approximately
30\% token retention.
mmVB uses combined single- and multi-image training.
HM is the harmonic mean of accuracy retention,
$\mathrm{acc}/\mathrm{acc}_{\mathrm{full}}$, and token compression,
$1-\mathrm{tok}/\mathrm{tok}_{\mathrm{full}}$.
Full Res.\ has an HM of $0.00$ because its token compression is zero.
Best and second-best results, excluding Full Res., are indicated by boldface
and underlining, respectively.
\end{minipage}

\caption{
Comparison of predictive performance, visual-token efficiency, and
computational efficiency across eight multi-image benchmarks.
}
\label{tab:resolution_efficiency_baseline_comp}

\end{table*}

\paragraph{Dataset.}
The single-image training set contains 484k samples from 30k images drawn from VQAv2, TextVQA, and GQA. Pairing each image with multiple queries provides a contrastive signal, as its optimal token level may vary by question, encouraging the predictor to model query$\times$image interactions rather than memorize image-specific levels. We also generate 50k multi-image samples from Mantis-Instruct~\cite{jiang2024mantis} and train with a curriculum that introduces multi-image data after single-image training. Mantis-Eval is a separate 217-sample benchmark released by the same authors.

\section{Experiments}

\noindent\textbf{Benchmarks and Settings.}
We evaluate on eight multi-image reasoning benchmarks spanning diverse task types and image counts: MuirBench~\cite{wang2025muirbench} (2.6k, 2--7 imgs), MIBench~\cite{liu2024mibench} (5k, 2--14 imgs), Mantis-Eval~\cite{jiang2024mantis} (217, 2--5 imgs), NLVR2~\cite{suhr2019corpus} (7k, 2 imgs), Visual Haystacks~\cite{wu2025visual} (20k, up to 10 imgs), MMIU~\cite{meng2025mmiu} (93k, 1--62 imgs), MIRB~\cite{zhao2024benchmarking} (1.1k, 2--6 imgs), and ReMI~\cite{kazemi2024remi} (2.6k, 2--6 imgs). We use Qwen3.5-4B as the base MLLM with greedy decoding and thinking mode disabled. Full dataset descriptions and training details are provided in the supplementary material.

\begin{figure}[!t]
    \centering
    \includegraphics[width=\linewidth]{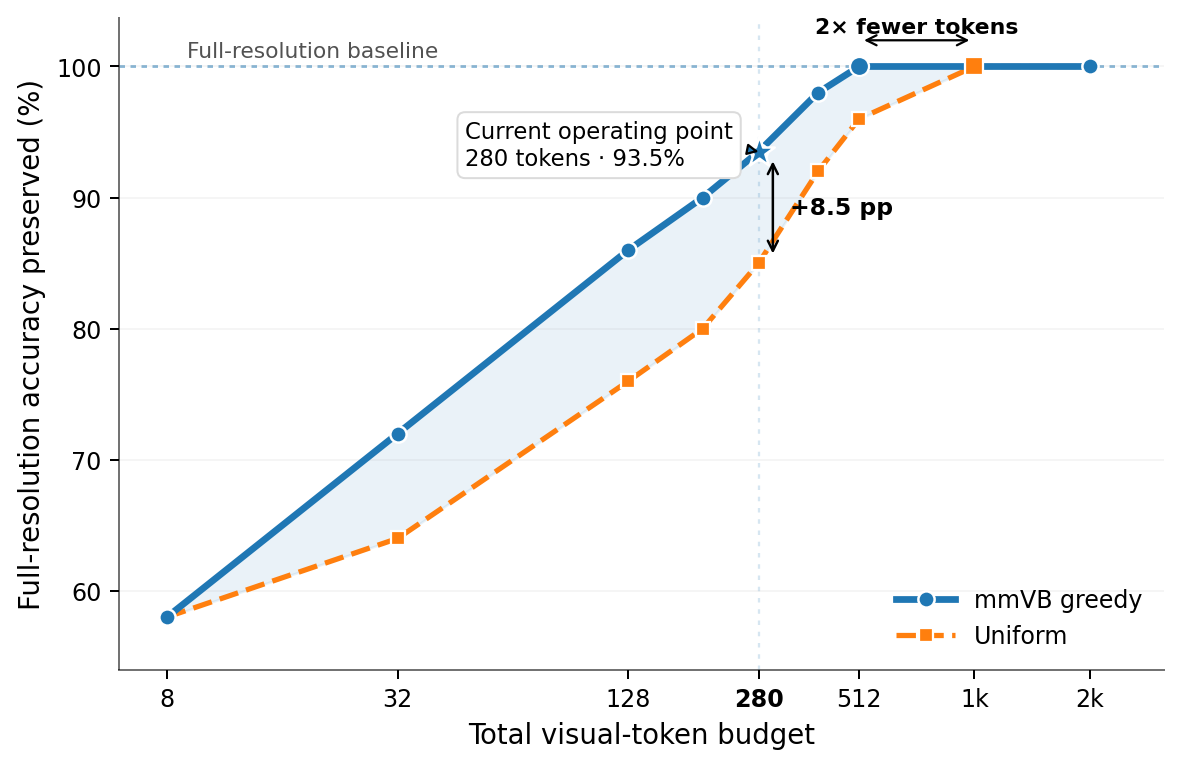}
    \caption{\textbf{Budget-constrained visual-token allocation.} Accuracy retention (mean across all eight benchmarks) as a function of average visual tokens per image. At matched budgets, query-conditioned greedy allocation consistently outperforms uniform allocation.}
    \label{fig:budget}
\end{figure}

\noindent\textbf{Baselines and Ablations.}
We compare against four post-encoding token-reduction baselines: \textit{VisionZip}~\cite{visionzip} (attention-based, 65\% dominant + 5\% contextual tokens), \textit{ToFu} \cite{pippi2025tofu} (dynamic token fusion via similarity merging), \textit{VisionTrim} \cite{yu2026visiontrim} (dominant selection + text-guided complement, retaining $\sim$30\%), and \textit{ZooPrune}~\cite{kim2026zoo} (zeroth-order gradient token pruning with keep ratio 0.11). All operate after the full ViT forward pass and cannot reduce encoder compute. We also report \textit{Full Res.} (upper bound), \textit{Uniform} (all images at 512$\times$512), \textit{Lower Bound} (all images at 32$\times$32), \textit{Random} (levels sampled from GT distribution), and \textit{mmVB$_{\text{SI}}$} (our predictor trained on single-image VQA only, to quantify the benefit of multi-image training data).\\
\noindent \textbf{Main Results.}
Table~\ref{tab:resolution_efficiency_baseline_comp} compares mmVB against post-encoding baselines. mmVB achieves the highest harmonic mean on six of eight benchmarks (average HM of 0.90), retaining 94.4\% of full-resolution accuracy while reducing visual tokens by 9.2$\times$ and prefill FLOPs by 7.3$\times$ (geometric mean across benchmarks). ZooPrune attains the best HM on MIBench and Visual Haystacks, where its aggressive token pruning (keep ratio 0.11) yields strong accuracy--compression trade-offs; however, it still requires a full ViT forward pass and cannot reduce encoder compute. VisionTrim is the strongest post-encoding baseline in accuracy, matching or exceeding mmVB on MIBench and MIRB, but it consumes 1.5--9.1$\times$ more tokens per sample than mmVB to achieve comparable accuracy. Even on benchmarks where mmVB trails VisionTrim by 1--4 points in accuracy, mmVB uses 4--9$\times$ fewer tokens and 4--7$\times$ fewer prefill FLOPs, yielding substantially higher HM. VisionZip achieves strong compression on few-image benchmarks (e.g., 8.3$\times$ on NLVR2) but retains most tokens on multi-image inputs (1.2$\times$ on Visual Haystacks), yielding inconsistent efficiency. ToFu collapses on benchmarks requiring cross-image reasoning (Visual Haystacks, MMIU) where query-unaware merging discards discriminative content. Critically, all post-encoding methods pay full ViT cost regardless of predicted importance; mmVB's pre-encoding resolution selection saves both encoder and LLM compute, aligning with findings that resolution reduction outperforms post-encoding compression at matched budgets~\cite{liao2026we}.\\
\noindent \textbf{Ablation Study.}
Table~\ref{tab:resolution_efficiency} isolates each component's contribution. The Random baseline samples levels from the ground-truth distribution but assigns them without regard to image content or query; it uses fewer tokens than mmVB on every benchmark because it randomly under-allocates to query-relevant images, but this results in substantially lower accuracy (e.g., 58.3\% vs.\ 62.9\% on MuirBench, 70.0\% vs.\ 82.8\% on NLVR2). The accuracy gap confirms that mmVB's higher token usage reflects meaningful allocation to images that need it, not inefficiency. For a controlled comparison at matched budgets, see Figure~\ref{fig:budget}. Adding multi-image training data (MI vs.\ SI) improves accuracy on all benchmarks except ReMI, with the largest gains on Visual Haystacks (+5.7pp) and NLVR2 (+3.6pp), where cross-image reasoning is critical.\\
\noindent \textbf{mmVB for Budget-Constrained Token Allocation.}
The default mmVB pipeline predicts the minimum token level for each image independently. In deployment, however, the total visual-token budget may be fixed by latency, memory, throughput, or cost constraints. We therefore extend mmVB to distribute an arbitrary total budget across all input images using its ordinal predictions. Let $b_i \in {b_0,b_1,b_2,b_3,b_4}$ denote the token allocation for image $I_i$, where each $b_k$ corresponds to one of five resolution levels. Given a total budget $B$, the allocation must satisfy
\begin{equation}
\sum_{i=1}^{N} b_i \leq B.
\end{equation}
Each promotion of image $I_i$ from level $k{-}1$ to level $k$ has utility $u_{i,k}=\sigma(z_{i,k})$ and token cost $\Delta b_k = b_k - b_{k-1}$, where $\sigma(z_{i,k})$ is the ordinal probability from the resolution head. Starting from the lowest level for all images, we greedily select the promotion with the highest $u_{i,k}/\Delta b_k$ and repeat until the budget is exhausted. This training-free procedure approximates a multiple-choice knapsack solution, prioritizing query-relevant images while leaving less relevant images at lower resolutions. As shown in Figure~\ref{fig:budget}, this query-conditioned allocation consistently outperforms uniform allocation under the same total budget and can recover full accuracy with substantially fewer tokens. These results show that visual-token efficiency depends not only on how many tokens are retained, but also on how they are distributed across images.

\paragraph{Black-box transfer.}
To test whether mmVB's learned resolution decisions reflect task-relevant visual information needs rather than model-specific artifacts, we evaluate with three black-box models that mmVB has never seen during training: Qwen3.5-27B, Claude Opus 4.6, and Pixtral Large. Images are resized according to mmVB's predicted levels (trained solely on Qwen3.5-4B), then passed directly to each model's API with no adaptation or fine-tuning. Qwen3.5-27B preserves 92.9\% accuracy with 9.1$\times$ visual token reduction; Claude Opus 4.6 preserves 89.8\% with 3.9$\times$ input token reduction; Pixtral Large preserves 89.9\% with 4.8$\times$ input token reduction (geometric means across benchmarks). This demonstrates that the resolution decisions generalize across architectures and scales, suggesting mmVB captures intrinsic image--query complexity rather than overfitting to the training model's visual encoder. We note that these results represent a lower bound on achievable performance: training a model-specific predictor (e.g., distilling from Qwen3.5-27B's own responses) would likely yield tighter accuracy retention, as the predictor could learn that model's particular sensitivity to resolution changes.

\begin{figure}[!t]
    \centering
    \includegraphics[width=\linewidth]{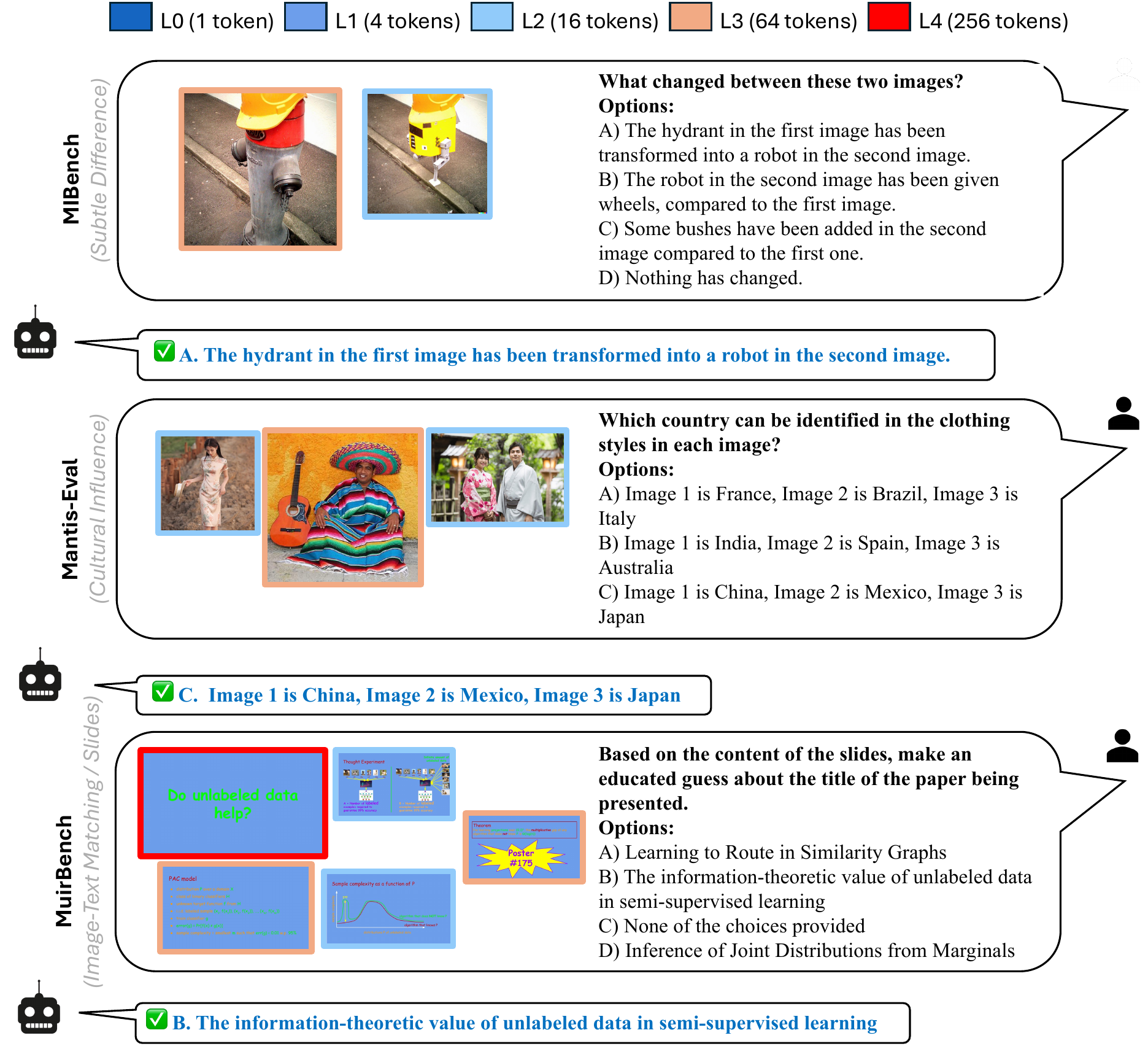}
    \caption{\textbf{Qualitative Examples from the benchmark.}}
    \label{fig:quals}
\end{figure}

\begin{table}[t]
\centering
\resizebox{\columnwidth}{!}{%
\begin{tabular}{@{}lrrrrrrrrrrrr@{}}
\toprule
& \multicolumn{4}{c}{\textbf{Qwen3.5-27B}} & \multicolumn{4}{c}{\textbf{Claude Opus 4.6}} & \multicolumn{4}{c}{\textbf{Pixtral Large}} \\
\cmidrule(lr){2-5} \cmidrule(lr){6-9} \cmidrule(lr){10-13}
\textbf{Benchmark} & \multicolumn{2}{c}{Acc (\%)} & \multicolumn{2}{c}{Tok/S} & \multicolumn{2}{c}{Acc (\%)} & \multicolumn{2}{c}{Tok/S} & \multicolumn{2}{c}{Acc (\%)} & \multicolumn{2}{c}{Tok/S} \\
\cmidrule(lr){2-3} \cmidrule(lr){4-5} \cmidrule(lr){6-7} \cmidrule(lr){8-9} \cmidrule(lr){10-11} \cmidrule(lr){12-13}
& Full & mmVB & Full & mmVB & Full & mmVB & Full & mmVB & Full & mmVB & Full & mmVB \\
\midrule
MuirBench       & 72.7 & 68.3 & 3,048 & 256 & 69.6 & 64.5 & 3,074 & 519 & 45.0 & 42.3 & 3,891 & 905 \\
MIBench         & 80.8 & 78.7 & 1,611 & 211 & 77.1 & 70.5 & 1,617 & 547 & 67.5 & 58.2 & 2,048 & 846 \\
Mantis-Eval     & 82.0 & 82.0 & 2,119 & 270 & 86.1 & 78.7 & 2,581 & 298 & 81.0 & 79.2 & 2,488 & 374 \\
NLVR2           & 93.8 & 88.8 & 1,649 & 70 & 91.2 & 86.1 & 1,368 & 250 & 83.7 & 78.6 & 2,048 & 281 \\
Vis.\ Haystacks & 77.6 & 68.8 & 2,187 & 358 & 71.6 & 69.3 & 3,586 & 1,908 & 78.9 & 77.4 & 7,987 & 2,392 \\
MMIU            & 66.9 & 61.7 & 2,950 & 360 & 63.9 & 64.8 & 3,787 & 1,314 & 53.2 & 49.3 & 4,096 & 944 \\
MIRB            & 70.7 & 70.5 & 1,794 & 690 & 76.2 & 62.8 & 2,584 & 951 & 66.5 & 64.1 & 4,966 & 668 \\
ReMI            & 23.0 & 18.0 & 2,561 & 118 & 32.2 & 23.0 & 1,472 & 356 & 27.7 & 18.1 & 2,417 & 474 \\
\bottomrule
\end{tabular}%
}
\caption{Black-box transfer: mmVB predictions (trained on Qwen3.5-4B) applied to unseen models without retraining.}
\label{tab:blackbox}
\end{table}
\paragraph{Does the mmVB learn per-image relevance?}
Yes. By assigning the lowest token budget to images irrelevant to the query, the predictor implicitly learns a relevance signal. On Visual Haystacks, a needle-in-haystack benchmark where only 1 of up to 10 images contains the answer mmVB assigns Level 0 to 87\% of images and reserves higher levels for the target. This emergent behavior arises without explicit relevance supervision: the GT generation procedure finds that irrelevant images can be maximally compressed without affecting correctness, so the predictor learns to map ``irrelevant to query'' directly onto minimal resolution. Examples are illustrated in Figure \ref{fig:quals}.

\begin{table*}[t]
\centering
\footnotesize
\resizebox{\textwidth}{!}{%
\begin{tabular}{@{}lrrrrrrrrrrrrrrrrrrr@{}}
\toprule
& & \multicolumn{6}{c}{\textbf{Accuracy (\%) $\uparrow$}} & \multicolumn{6}{c}{\textbf{Vis Tok/Samp $\downarrow$}} & \multicolumn{6}{c}{\textbf{TTFT (s) $\downarrow$}} \\
\cmidrule(lr){3-8} \cmidrule(lr){9-14} \cmidrule(lr){15-20}
\textbf{Benchmark} & \textbf{N} & Full & Uni & Low & Rand & SI & MI & Full & Uni & Low & Rand & SI & MI & Full & Uni & Low & Rand & SI & MI \\
\midrule
Muir & 2,600 & \gray{63.6} & 63.8 & \dgray{52.7} & 58.3 & 61.7 & \textbf{62.9} & \gray{3,236} & 1,109 & \dgray{5} & \textbf{370} & 388 & 389 & \gray{3.5} & 0.6 & \dgray{0.6} & 0.7 & 0.8 & \textbf{0.5} \\
MIB & 5,000 & \gray{78.1} & 77.4 & \dgray{64.7} & 72.6 & 74.6 & \textbf{77.9} & \gray{1,237} & 1,178 & \dgray{5} & \textbf{125} & 231 & 248 & \gray{1.6} & 1.0 & \dgray{0.5} & \textbf{0.5} & 0.7 & 0.7 \\
Mantis & 217 & \gray{69.9} & 69.1 & \dgray{49.3} & 59.9 & 65.4 & \textbf{68.2} & \gray{2,119} & 637 & \dgray{3} & \textbf{73} & 103 & 118 & \gray{2.2} & 0.9 & \dgray{0.5} & \textbf{0.6} & \textbf{0.6} & \textbf{0.6} \\
NLVR2 & 6,967 & \gray{86.5} & 86.9 & \dgray{56.8} & 70.0 & 79.2 & \textbf{82.8} & \gray{1,649} & 512 & \dgray{2} & \textbf{63} & 75 & 89 & \gray{2.8} & 0.6 & \dgray{0.5} & \textbf{0.5} & \textbf{0.5} & 0.6 \\
VHay & 20,100 & \gray{69.5} & 69.5 & \dgray{51.7} & 58.5 & 63.9 & \textbf{69.6} & \gray{2,187} & 2,090 & \dgray{8} & \textbf{271} & 571 & 594 & \gray{11.8} & 0.7 & \dgray{1.2} & 4.0 & 6.0 & \textbf{0.6} \\
MMIU & 93,584 & \gray{50.9} & 58.7 & \dgray{37.3} & 46.4 & 50.1 & \textbf{53.2} & \gray{2,980} & 1,735 & \dgray{7} & \textbf{203} & 217 & 453 & \gray{7.9} & 0.7 & \dgray{2.1} & \textbf{0.6} & 0.6 & 0.6 \\
MIRB & 1,124 & \gray{67.2} & 63.0 & \dgray{40.2} & 45.7 & 53.3 & \textbf{56.5} & \gray{1,756} & 791 & \dgray{3} & \textbf{123} & 182 & 193 & \gray{2.5} & 1.5 & \dgray{1.1} & \textbf{1.1} & 1.2 & 1.3 \\
ReMI & 2,600 & \gray{24.7} & 25.7 & \dgray{17.1} & 18.4 & \textbf{20.0} & 19.2 & \gray{2,573} & 638 & \dgray{3} & \textbf{87} & 143 & 155 & \gray{22.2} & 16.4 & \dgray{15.7} & \textbf{14.2} & \textbf{14.2} & 15.2 \\
\bottomrule
\end{tabular}%
}
\caption{Ablation: accuracy and efficiency across eight multi-image benchmarks. Full = native resolution (reference). Uni = all images at 512$\times$512. Low = all images at 32$\times$32 (lower bound). Rand = random level (distribution-matched). SI = mmVB single-image training. MI = mmVB single+multi-image training. \textbf{Bold} = best (excl.\ Full/Low).}
\label{tab:resolution_efficiency}
\end{table*}

\section{Related Work}

Our work relates to visual token compression, visual resampling, multi-image reasoning, and efficient attention. Prior methods reduce visual-token cost through pruning, merging, or compact representations \cite{dynamicvit,tome,fastv,visionzip,sparsevlm,llavaprunerge,minigemini,tokenpacker,m3,xing2025pyramiddropacceleratinglargevisionlanguage,hu2026illava,shen2024longvuspatiotemporaladaptivecompression,han2025filtercorrelatecompresstrainingfree,renggli2022learning}. However, recent work suggests that simple input downsampling can outperform several post-encoding compression methods at matched compression ratios \cite{liao2026we}. Most approaches also process each image independently, rather than allocating a shared, question-conditioned token budget across multiple images. Latent resampling methods such as BLIP-2, Perceiver IO, and Flamingo map visual inputs into compact token sets \cite{blip2,perceiverio,flamingo}, but typically use a fixed budget per image. Multi-image models, including Mantis, LLaVA-NeXT-Interleave, LLaVA-OneVision, LongVA, and MuRAG, improve reasoning over multiple images or long multimodal contexts \cite{jiang2024mantis,llavanextinterleave,llavaonevision,murag}, but do not explicitly adapt token allocation to question relevance. AVAM~\cite{zeng2025avam} introduces query-conditioned visual anchoring for multi-image QA, but targets accuracy improvement rather than efficiency: its collaborative decoding mechanism runs both global and compressed visual branches, increasing inference cost. In contrast, mmVB operates before encoding and reduces both vision-encoder and LLM compute in a single forward pass. MIRAGE~\cite{wu2025visual} uses visual retrieval to select which images to process, scaling to thousands of images by discarding irrelevant ones entirely; mmVB instead retains all images but allocates different token budgets, which is necessary for tasks where every image may contribute (e.g., comparison, counting). Efficient attention methods such as FlashAttention and FlashAttention-2 reduce long-context memory and computation \cite{flashattention,flashattention2}, although they make attention-map-based pruning less practical. Overall, existing methods often use fixed budgets, operate on images independently, require attention-score access, or train new visual-language interfaces \cite{lee2026frequency,kim2026zoo,alvar2025divprune,kimhi2025cares,zhang2024synergistic,ye2025infotok,lee2024video,choudhury2024don,wei2023joint,zhoustatistically,xie2026saliency,rubab2026dyna,ding2023prune,chen2023efficient,ryoo2021tokenlearner,lei2025rethinking,eliopoulos2025pruning}. In contrast, we introduce lightweight, plug-and-play visual budgeting for multi-image reasoning that is question-conditioned, requires no base-model retraining, and avoids dense attention-map access at inference time.
\section{Conclusion}
We presented mmVB, a plug-and-play module that assigns query-aware token budgets before visual encoding. It achieves 9$\times$ token reduction, 7$\times$ fewer prefill FLOPs, and 5$\times$ faster TTFT while retaining 94.4\% accuracy, with negligible latency and strong cross-model transfer.

\begin{figure}[!t]
    \centering
    \includegraphics[width=\linewidth]{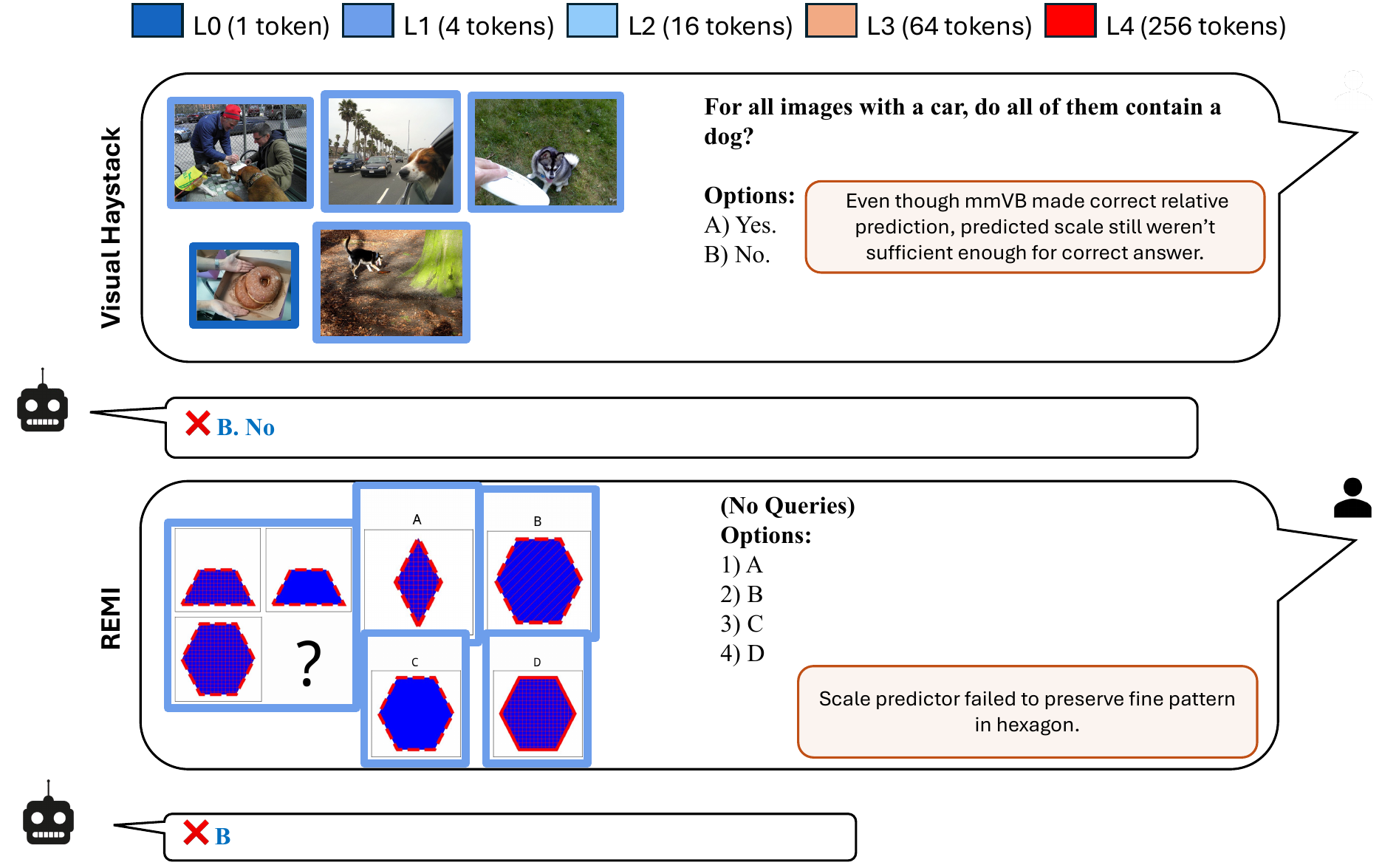}
    \caption{\textbf{Limitations.} Scale predictor may underestimate tokens needed to answer questions. May also fail when there's no text query.}
    \label{fig:failures}
\end{figure}

\section{Limitations}
The scale predictor is trained exclusively on natural-image QA datasets (VQAv2, TextVQA, GQA). Domains with fundamentally different resolution requirements, such as medical imaging where diagnostically relevant features may occupy only a few pixels within a high-resolution scan, are not represented in the training distribution. In such domains, the predictor may under-allocate tokens because it has not learned that subtle fine-grained features (e.g., lesion boundaries, microcalcifications) require full resolution regardless of query type. See Figure \ref{fig:failures} for examples.
\bibliography{custom}
\appendix
\section{Architecture Details}
\label{app:arch}

\paragraph{Image Encoding.}
Each image is encoded by a frozen ImageNet-pretrained EfficientNet-B0 backbone~\cite{tan2019efficientnet}. Its 1280-dimensional feature is projected to the shared dimension $d=256$ using a linear layer and dropout. Processing low-resolution thumbnails with EfficientNet-B0 is substantially cheaper than applying the MLLM vision encoder to every image.

\paragraph{Text Encoding.}
The query $Q$ is tokenized and passed through the frozen MLLM embedding layer. The resulting $d_{\text{LLM}}$-dimensional token representations are mean-pooled and projected to the shared space by $\operatorname{Linear}(d_{\text{LLM}},64)$, ReLU, $\operatorname{Dropout}(0.3)$, $\operatorname{Linear}(64,256)$, and LayerNorm. The narrow bottleneck limits memorization and encourages the retention of task-relevant query information.

\paragraph{Context Fusion.}
The query embedding and $N$ image embeddings form a sequence of $N+1$ tokens with dimension $256$. A two-layer Transformer encoder with four attention heads and a feed-forward dimension of $512$ performs query-conditioned and cross-image reasoning, allowing each image to estimate both its required visual detail and its relative importance.

\paragraph{Resolution Head.}
Each image representation is passed through an MLP with dimensions $256 \rightarrow 64 \rightarrow 4$. The four outputs define cumulative ordinal-regression thresholds
\[
P(\ell_i \geq k)=\sigma(z_{i,k}), \qquad k \in \{1,2,3,4\},
\]
and the predicted level is
\[
\hat{\ell}_i=\sum_{k=1}^{4}\mathbb{I}\left[\sigma(z_{i,k})>0.5\right].
\]
The four thresholds are trained independently without an explicit monotonicity constraint. While independent sigmoids can produce non-monotone probabilities, the thresholding rule above always yields a valid level; in practice, violations are infrequent because the ordinal loss encourages consistent threshold ordering.

\paragraph{Inference and Parameter Count.}
Each image is resized according to its predicted token level, with the target resolution capped at the original dimensions to avoid upscaling. The budgeter processes only $N+1$ tokens through two Transformer layers, adding approximately $2\mathrm{ms}$ of overhead. It contains about $1.6$ million trainable parameters: 328k in the image projection, 181k in the text encoder, 1,054k in the Transformer, and 17k in the resolution head. The frozen EfficientNet-B0 adds 5.3 million parameters. The MLLM embedding layer must also remain in memory, but introduces no additional cost when co-deployed with the target MLLM.

\paragraph{Design Choices.}
We select EfficientNet-B0 as the frozen image encoder because it offers the best accuracy among models at the $\sim$5M parameter scale (77.1\% ImageNet top-1), outperforming MobileNetV2 (72.0\%), MobileNetV3-Large (75.2\%), and DeiT-Tiny (72.2\%). TinyViT-5M achieves higher accuracy (79.1\%) but requires 3.3$\times$ more compute; since the encoder processes $N$ thumbnails per sample, this overhead compounds. Because the backbone is frozen, parameter count is immaterial; only representation quality matters for downstream level prediction. The remaining hyperparameters were selected via validation performance on the token-based training set (Table~\ref{tab:arch_ablation}). Two Transformer layers provide sufficient capacity for cross-image reasoning without overfitting; a single layer under-performs on multi-image benchmarks where relative importance must be inferred. Five ordinal levels match the natural 4$\times$ token gaps in vision encoders, providing meaningful resolution distinctions at each step. The 64-dim text bottleneck restricts information flow from the 2560-dim LLM embeddings; wider bottlenecks memorize per-sample query identity while narrower ones discard task-type signal.

\begin{table}[h]
\centering
\footnotesize
\setlength{\tabcolsep}{4pt}
\begin{tabularx}{\linewidth}{@{}lXcc@{}}
\toprule
\textbf{Component} & \textbf{Variant} & \textbf{bal\_acc} & \textbf{$\pm$1 acc} \\
\midrule
\multirow{3}{*}{Levels}
& 3 levels & 65 & 79 \\
& 5 levels (ours) & 71 & 86 \\
& 7 levels & 70 & 86 \\
\midrule
\multirow{3}{*}{Text bottleneck}
& None (direct 2560$\to$256) & 55 & 71 \\
& 64-dim (ours) & 71 & 86 \\
& 16-dim & 50 & 71 \\
\midrule
\multirow{2}{*}{Cross-image}
& Text-only (no Transformer) & 51 & 55 \\
& 2-layer Transformer (ours) & 71 & 86 \\
\bottomrule
\end{tabularx}
\caption{Architecture ablation on the validation set. 
bal\_acc = balanced accuracy across levels; $\pm$1 = fraction of predictions 
within one level of ground truth.}
\label{tab:arch_ablation}
\end{table}

\section{Additional Results}

\subsection{Per-task performance breakdown}

\begin{figure*}[t]
    \centering
    \includegraphics[width=\textwidth]{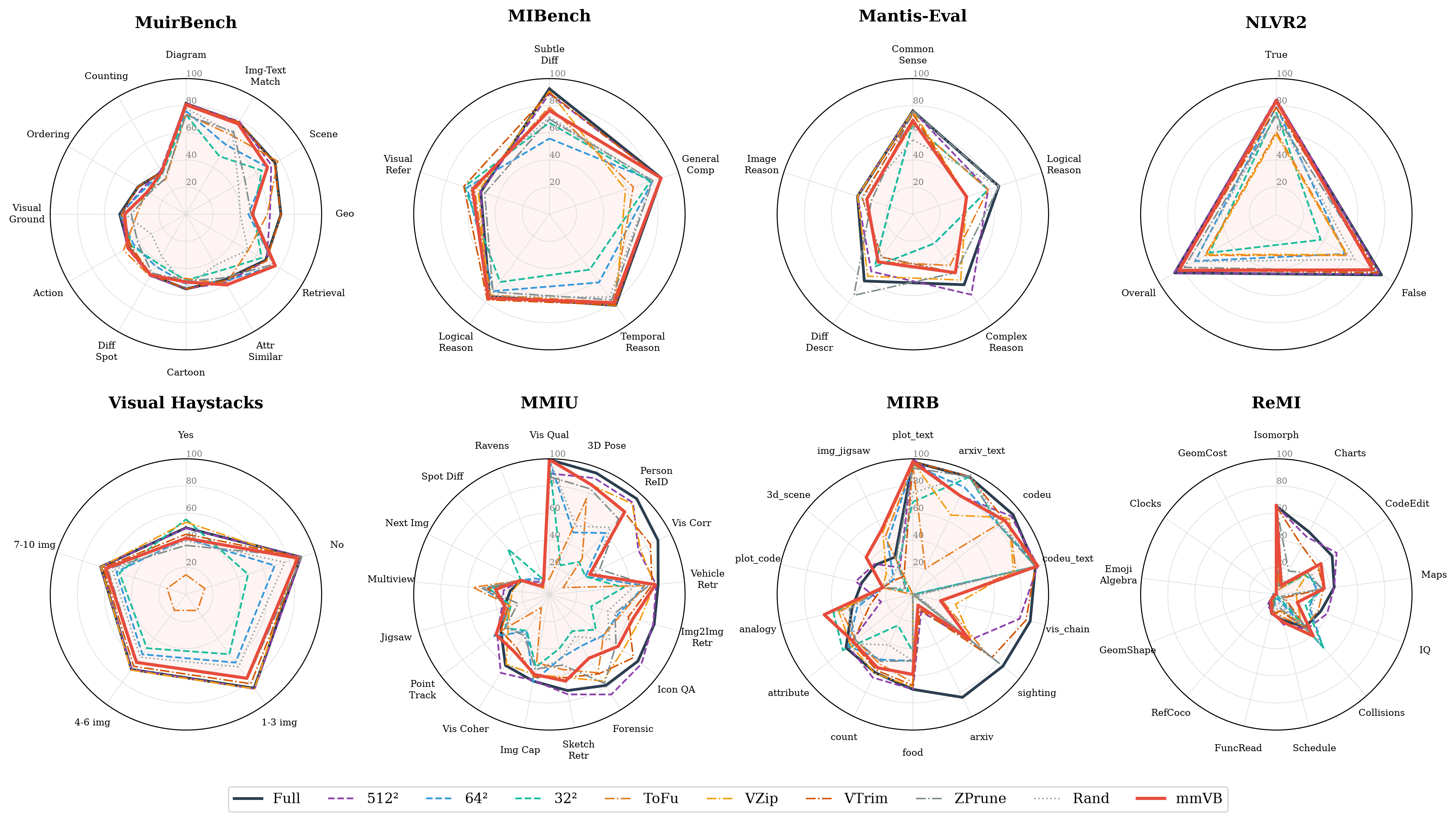}
    \caption{Per-task accuracy across eight benchmarks. Each subplot represents one benchmark. Solid lines show Full-res and mmVB; dashed lines show uniform resolutions; dash-dot lines show post-encoding baselines; and dotted lines show the distribution-matched random baseline. mmVB closely tracks Full-res while using 9.2$\times$ fewer visual tokens (geometric mean across benchmarks).}
    \label{fig}
\end{figure*}

Figure~\ref{fig} shows that mmVB closely follows Full-res across most tasks for every benchmarks. Uniform downsampling causes uneven degradation, with the largest drops on detail-sensitive tasks such as Charts, 3D Pose, Person ReID, and visual chain reasoning. Coarser tasks remain relatively stable even at lower resolutions. Post-encoding baselines show greater task-level variation and occasional failures, while ReMI remains challenging for all methods, especially on numerical reasoning tasks. Overall, mmVB preserves most of the Full-res performance while using substantially fewer visual tokens.

\subsection{mmVB performance on Single Image} 
In this section, we evaluate mmVB on single-image tasks. Since many standard benchmarks, such as VQAv2 and TextVQA, focus on general visual question answering, we select MM-Vet and POPE to assess transferability to single-image settings. MM-Vet evaluates the integration of core vision-language capabilities, including recognition, OCR, knowledge, language generation, spatial reasoning, and mathematics. Table~\ref{tab:mmvet} reports results on MM-Vet~\cite{yu2023mmvet}, a single-image open-ended VQA benchmark (218 samples) scored by an LLM judge.

\begin{table}[t]
\centering
\begin{tabular}{@{}lrr@{}}
\toprule
\textbf{Method} & \textbf{Score (\%)} & \textbf{Vis tok/samp} \\
\midrule
\rowcolor{fullresgray}
Full Res. & 59.8 & 1,146 \\
mmVB  & 53.3 & 63 \\
\bottomrule
\end{tabular}
\caption{MM-Vet results (Qwen3.5-4B, Claude judge)}
\label{tab:mmvet}
\end{table}

\section{Dataset Details}

\begin{table*}[t]
\centering
\resizebox{\textwidth}{!}{%
\begin{tabular}{lcccccccc}
\toprule
& \textbf{MuirBench}
& \textbf{MIBench}
& \textbf{Mantis}
& \textbf{NLVR2}
& \textbf{VHs}
& \textbf{MMIU}
& \textbf{MIRB}
& \textbf{ReMI} \\
\midrule
Samples
& 2,600 & 5,000 & 217 & 6,967
& 20,100 & 93,584 & 1,124 & 2,600 \\
Images/sample
& 2--7 & 2--14 & 2--5 & 2
& 1--10 & 1--62 & 2--6 & 2--6 \\
Tasks
& 12 & 5 & 6+ & 1
& 1 & 52 & 14 & 13 \\
Format
& MC & MC & MC+SA & T/F
& Y/N & MC & Mixed & Free-form \\
\bottomrule
\end{tabular}%
}
\caption{Overview of the evaluation benchmarks. MC = multiple-choice, SA = short-answer, T/F = true/false, and Y/N = yes/no. Mixed includes multiple-choice, yes/no, counting, and free-form answers. VHs denotes Visual Haystacks.}
\label{tab:benchmarks}
\end{table*}

\subsection{Benchmark details}

We evaluate multi-image reasoning on eight complementary benchmarks spanning diverse tasks, image counts, and reasoning demands. Table~\ref{tab:benchmarks} summarizes the eight benchmarks.

\subsection{MuirBench}

MuirBench \cite{wang2025muirbench} contains 11,264 images and 2,600 multiple-choice questions across 12 tasks. Each question requires joint reasoning over 2--7 images and covers relations such as narrative, temporal, complementary, cropped or zoomed, and multi-view scenes. Tasks include geographic and diagram understanding, retrieval, counting, difference spotting, and action understanding. The benchmark tests whether models reason across images rather than rely on single-image shortcuts.

\subsection{MIBench}

MIBench \cite{liu2024mibench} contains 5,000 multiple-choice questions across five categories: comparison, subtle difference detection, temporal reasoning, logical reasoning, and visual referring. Samples use text-rich documents, natural photographs, and synthetic scenes. The benchmark evaluates whether multi-image models can follow varied instructions over image sets.

\subsection{Mantis-Eval}

Mantis-Eval \cite{jiang2024mantis} contains 217 human-annotated problems in multiple-choice and short-answer formats. Each sample includes 2--5 images and covers common-sense, complex, logical, and fine-grained visual reasoning. Despite its small size, the benchmark is difficult: GPT-4V achieves 62.67\%.

\subsection{NLVR2}

NLVR2 \cite{suhr2019corpus} evaluates reasoning over image pairs. Each sample presents two photographs and a natural-language statement that must be classified as \textit{True} or \textit{False}. The statements require reasoning about object properties, spatial relations, comparison, counting, and set-level quantification. The test set contains 6,967 samples, including 3,558 true and 3,409 false instances.

\subsection{Visual Haystacks}

Visual Haystacks \cite{wu2025visual} is a vision-centric Needle-In-A-Haystack benchmark for large multimodal models. Each sample mixes a small set of ``needle'' images with distractors and asks whether all or any needle images also contain a target object. The full dataset contains 20,100 samples with 1 to more than 10,000 images. We cap each haystack at 10 images by retaining all positives and randomly sampling negatives. The benchmark tests retrieval and reasoning over large image sets.

\subsection{MMIU}

MMIU \cite{meng2025mmiu} contains 93,584 samples, 77,659 images, and about 52 tasks spanning semantic, temporal, and spatial relations. Tasks include forensic detection, correspondence, temporal ordering, action assessment, jigsaw solving, 3D pose estimation, and Raven's matrices. Each sample contains 1--62 images and a multiple-choice question with up to 10 options. The random baseline is 27.4\%, and the best reported model, GPT-4o, achieves 55.7\%.

\subsection{MIRB}

MIRB \cite{zhao2024benchmarking} contains 1,124 samples across 14 tasks covering perception, knowledge, reasoning, and multi-hop reasoning. Tasks include visual analogy, cross-image counting, multi-view 3D understanding, code and paper comprehension, ingredient analysis, jigsaw puzzles, plot interpretation, landmark recognition, and visual chain reasoning. Answer formats include multiple-choice, yes/no, counting, and free-form extraction.

\subsection{ReMI}

ReMI \cite{kazemi2024remi} contains 2,600 samples across 13 mathematical, spatial, temporal, and logical reasoning tasks. These include Charts, Clocks, CodeEdit, Collisions, EmojiAlgebra, FuncRead, GeomCost, GeomShape, IQ, Isomorphism, Maps, RefCoco, and Schedule. Each sample contains 2--6 images and requires a short free-form answer. ReMI emphasizes quantitative and structured reasoning rather than visual matching alone.

\subsection{mmVB Training Data Composition}

We generate resolution labels over a diverse training set combining single-image and multi-image sources:

\noindent \textbf{Single-image} (484k samples): VQAv2, TextVQA, GQA covering natural scenes, text-heavy images, and compositional reasoning respectively.\\
\noindent \textbf{Multi-image} (50k samples from Mantis-Instruct): 17 subsets including NLVR2, LLaVA-multi, spot-the-diff, visual storytelling, coinstruct, and others, covering 2--10+ images per sample with diverse multi-image reasoning tasks.

Samples where the MLLM answers incorrectly at full resolution are discarded, as they provide no valid signal for minimum-resolution search. Generation is parallelized across 8 GPUs with manifest sharding for throughput.
\end{document}